\documentclass{article}

\usepackage{arxiv}
\usepackage[utf8]{inputenc} 
\usepackage[T1]{fontenc}    
\usepackage{hyperref}       
\usepackage{url}            
\usepackage{booktabs}       
\usepackage{amsfonts}       
\usepackage{nicefrac}       
\usepackage{microtype}      
\usepackage{lipsum}
\usepackage{graphicx}

\title{Trust, but verify}

\author{
 Michael J. Yuan \\
  ByteTrade Labs \\
  Austin, Texas, USA \\
  \texttt{michael@michaelyuan.com} \\
   \And
 Carlos Lospoy \\
  Nolais \& Co/Labs \\
  New York, New York, USA \\
  \texttt{carlos@lospoy.com}
  \And
 Sydney Lai \\
  Gaia Foundation \\
  New York, New York, USA \\
  \texttt{sydney@gaianet.ai} \\
  \AND
  James Snewin \\
  Eigen Labs \\
  Brisbane, Australia \\
  \texttt{james.snewin@eigenlabs.org} \\
  \And
  Ju Long \\
  McCoy College of Business\\
  Texas State University \\
  San Marcos, Texas, USA \\
  \texttt{jl38@txstate.edu} \\
}

\begin{document}
\maketitle

\begin{abstract}
Decentralized AI agent networks, such as Gaia, allows individuals to run customized LLMs on their own computers and then provide services to the public. However, in order to maintain service quality, the network must verify that individual nodes are running their designated LLMs. In this paper, we demonstrate that in a cluster of mostly honest nodes, we can detect nodes that run unauthorized or incorrect LLM through social consensus of its peers. We will discuss the algorithm and experimental data from the Gaia network. We will also discuss the intersubjective validation system, implemented as an EigenLayer AVS to introduce financial incentives and penalties to encourage honest behavior from LLM nodes.
\end{abstract}


\section{Introduction}

Two years after OpenAI released ChatGPT, there is an explosion of open-source LLMs both in quality and quantity [1]. The best open-source LLMs are now matching the best closed-source LLMs in performance [2], while offering much more variety in terms of model size, speed, and specific "skills", such as math and coding [3]. The rise of open source enables individual users to run customized LLMs and agent applications on their own computers without centralized gatekeepers. The decentralized LLM inference has many benefits, such as privacy, low cost, speed, and availability. But, more importantly, it fosters an ecosystem of customized LLM services based on models fine-tuned and supplemented by proprietary data and knowledge.

Decentralized AI inference networks, such as Gaia [4], allow individuals (and businesses) to run AI models on their own computers. Each "Gaia node" runs a set of AI models, including LLMs, and a knowledge base. It advertises and makes available its services to the public. 

Gaia nodes can pool together their computing resources into a "Gaia domain", which is a set of Gaia nodes that are supposed to run the same LLM and knowledge base. A Gaia domain load-balances across its Gaia nodes to serve the in-demand customized LLMs to the public and share revenues with the model creators. A popular Gaia domain could have over 1,000 nodes at any given time. However, as those networks must stay permissionless, to reduce censorship, protect privacy and lower the barrier of participation, it is possible for some Gaia nodes to cheat and run a different LLM or knowledge base than the ones it's domain requires. The network must have a mechanism to automatically detect and penalize the nodes that cheat. 

\pagebreak

\section{Past research}

Researchers have indicated that they could deterministically verify any computing machine using cryptographic algorithms. 

Zero Knowledge Proof (ZKP) allows us to verify the outcome of a computation process without knowing the internals of the computing system [5]. In the context of LLM verification, we can build a ZKP circuit for an LLM, which represents the inference process. The LLM service provider (Gaia node) generates a proof for each incoming request (i.e., the LLM prompt) indicating that it has correctly executed the inference process [6]. Any external party (validator) can then check the proof and confirm that it is generated by the specific LLM without knowing the exact software setup on the machine.

However, ZKP also has some severe limitations that make it impossible to use in production environments. 

\begin{itemize}
\item The ZKP circuit must be generated for each LLM. That requires vast engineering efforts since there are tens of thousands of open-source LLMs on Huggingface alone. 
\item Even the state-of-the-art ZKP algorithms running on expensive Nvidia GPUs require 13 minutes to generate proof for a single inference execution for a small 13B LLM [6]. That is 2 orders of magnitude slower (and 2 orders of magnitude more resource consuming) than the inference task itself. Since LLM inference is already compute-bound, spending $100\times$ on compute just for verification is not feasible.
\item The memory cost of generating ZKP proofs is also staggering. A toy LLM with one million parameters would require over 25GB of RAM even with the most optimized algorithms [7]. Scaling that to LLMs of $10,000\times$ of that size would require computing resources far beyond any reasonable computer.
\item If the LLM itself is open source, it is possible to fake the proof, which would render the ZKP entirely useless in most decentralized AI inference systems where open-source is a requirement. 
\end{itemize}

A second cryptographic approach is to use hardware-based Trusted Execution Environment (TEE) that are built into CPUs and GPUs. TEE can generate signed hashes (i.e., attestation) for software and data on the processor. That ensures that the software and data exactly matches a specified version. Since the TEE is a hardware feature, it is impossible for the computer operator or owner to manipulate or fake a proof. 

However, TEE too, is not ready for large-scale AI inference. 

\begin{itemize}
\item The TEE reduces raw hardware (CPU) performance by a factor of up to $2\times$ [8], which is very hard to accept since LLM inference is already a compute-bound task.
\item There are very few GPUs or AI accelerators that support TEEs. The high-end H-100 is the first Nvidia GPU with TEE support [9].
\item Even if we can verify that a specific LLM is running inside a TEE, it is still impossible to verify that an LLM server is actually using that LLM to serve Internet requests. There are many moving parts in an AI server that resides outside of the TEE. 
\item Distributing private keys to decentralized and permissionless TEE devices is a major challenge that requires specialized software and operational procedures [10]. 
\end{itemize}

As we have seen above, the cryptographic approaches are slow, expensive, and generally impractical with todayâ€™s consumer-grade computer hardware. A more promising approach is to use cryptoeconomic mechanisms, which optimistically assume that most players in a decentralized inference network are honest and then use social consensus to detect dishonest players. Through staking and slashing, the network can incentivize honest players and punish dishonest players, and hence creating a virtuous cycle. 

Since the LLMs are non-deterministic, we need a group of validators that continuously sample the LLM service providers, and vote on the answers to detect dishonest LLM service providers and also dishonest validators. The sampling, data collection and voting are all conducted off-chain through interactions between the validators and nodes. But the results are recorded on-chain and have cryptoeconomic consequences. 

The validators could compare LLM outputs in the embedding space [11]. Each word token is mapped to a vector of real numbers in a high-dimensional space (e.g., 768 ot 1536 dimensions). The vector, known as an embedding, represents the initial (or most common) meaning of the word token. The embeddings in a sentence or paragraph are then sequentially transformed by a specially trained LLM to adjust for its context (e.g., the tokens before it) in the sentence. The last token has supposedly "seen" every word in the sentence, and hence its context-adjusted embedding is taken as the overall embedding of the sentence. The result is that sentences that have similar meanings will have embedding vectors that are close together. That allows the validators to compare the similarity of outputs from Gaia nodes.

EigenLayer's AVS [12] (Actively Validated Service or Autonomous Verifiable Service) provides a set of smart contracts that allows independent operators and validators to stake crypto assets, and vote on intersubjective results from the sampling. Once a vote reaches consensus, known as an intersubjective consensus [13], the smart contracts automatically execute to reward or punish the operators and validators.

The voting rules and algorithms are the subject of this research. In the rest of this paper, we will first present hypotheses on how to distinguish dishonest LLM service providers (i.e., Gaia nodes) from a group of honest providers. We will then run experiments and validate the hypotheses through empirical data. We will also discuss AVS implementation design and considerations for such a system.

\section{Hypotheses}

Recall that a Gaia node is a network server that runs a set of AI models and knowledge bases. It promises to run the AI model and the knowledge base it advertises on the Gaia network.

We hypothesize that by asking questions to a set of Gaia nodes, and observing the statistical distributions of the answers, we will be able to detect outliers that run different LLMs or knowledge bases than the majority.

Specifically, when we ask a question to all nodes in a Gaia domain, the answers will form a tight cluster in a high-dimensional embedding space, and if there are outliers that lies far outside of the cluster, the corresponding nodes are likely running a different LLM or knowledge base than the ones required for this domain. 

We formulate a set of questions $Q$ and a set of Gaia nodes $M$. In each experiment, we choose a question $q$ and send it to the Gaia node $m$ for an answer. 

\begin{equation}
q \in Q
\end{equation}

\begin{equation}
m \in M
\end{equation}

The question is repeated $n$ times for each node $m$ to create a distribution of the answers. We will hence have $n$ answers. Each answer is passed to an LLM embedding model to generate a $z$ dimensional vector point that is representative of its semantic meaning. In the case of the \texttt{gte-Qwen2-1.5B-instruct} embedding model, $z$ is $1536$. The $z$ dimensional vector point from the $i$th answer of question $q$ on node $m$ is written as follows. 

\begin{equation}
X_i(q,m) = [x_{0,i}(q,m), ... x_{z,i}(q,m)]
\end{equation}

We can compute the mean point of answers generated from question $q$ and node $m$ as follows.

\begin{equation}
\overline{X}(q,m) = \frac{\left[\sum_{i=0}^{n-1} x_{0,i}(q,m), ... \sum_{i=0}^{n-1} x_{z,i}(q,m)\right]}{n}
\end{equation}

For a given pair of $q,m$ and $q',m'$, the difference between the answers can be measured as the distance between the mean points. 

\begin{equation}
D(q, m, q', m') = \sqrt{\sum_{j=0}^{z-1} \left(\frac{\sum_{i=0}^{n-1} x_{j,i}(q,m) - \sum_{i=0}^{n-1} x_{j,i}(q',m')}{n}\right)^2}
\end{equation}

The consistency of the $n$ answers generated by a node $m$ for a question $q$ is measured as the standard deviation of the vector points.

\begin{equation}
\sigma_j(q,m) = \sqrt{\frac{\sum_{i=0}^{n-1} \left(x_{j,i} - \frac{\sum_{i=0}^{n-1} x_{j,i}}{n}\right)^2}{n}}
\end{equation}

We can compute the Root-Mean-Square (RMS) to represent the internal variance of the distribution. 

\begin{equation}
\overline{\sigma}(q,m) = \sqrt{\frac{1}{z}\sum_{j=0}^{z-1}\sigma_j^2(q,m)}
\end{equation}

Our hypotheses are therefore as follows. 

Hypothesis 1: When two Gaia nodes with different LLM or knowledge base configurations $m$ and $m'$ answer the same question $q$, the answers form distinct distributions that are specific to the Gaia node. This allows an outside observer to easily tell which Gaia node generated a given answer. 

\begin{equation}
D(q,m,q,m') > 3 \times (\overline{\sigma}(q,m) + \overline{\sigma}(q,m'))
\end{equation}

Hypothesis 2: When a Gaia node $m$ answers two independent questions $q$ and $q'$, the answers form distinct distributions that are specific to the question. This allows an outside observer to easily tell which question resulted a given answer. 

\begin{equation}
D(q,m,q',m) > 3 \times (\overline{\sigma}(q,m) + \overline{\sigma}(q',m))
\end{equation}

\section{Experiments}

A Gaia node has two key components that affect the responses that can be observed from an outside validator: the LLM or the knowledge base (vector database).

Hence we ran two sets of experiments to test the above hypotheses on LLM and knowledge base variations respectively. The source code and data to replicate all the analyses and results in this paper are available as an open source GitHub repo [14].

\subsection{Responses from different models}

In the first experiment, we aim to test whether we can reliably distinguish responses from different LLMs. We set up 3 Gaia nodes for the following 3 open source LLMs.

\begin{itemize}
\item Llama 3.1 8b by Meta AI [15].
\item Gemma 2 9b by Google [16].
\item Gemma 2 27b by Google [16].
\end{itemize}

Each model was queried with 20 factual questions covering topics such as science, history, and geography (see Appendix A). Each question was repeated 25 times per model, yielding 500 responses per model and 1,500 responses in total.

\subsection{Responses from different knowledge bases}

In the second experiment, we test whether we can reliably distinguish responses from different knowledge bases used by the same LLM.

We set up two Gaia nodes with the same LLM, Gemma-2-9b by Google [16]. The knowledge bases are vector databases created from text files [17]. In this experiment, we adopted knowledge bases from Paris and London Wikipedia pages respectively [18, 19]. The knowledge bases are created with the \texttt{nomic-embed-text-1.5} embedding model [20]. The vector database files are available as follows.

\begin{itemize}
\item Paris [21].
\item London [22].
\end{itemize}

Each knowledge node was queried with 20 factual questions with even coverage on Paris and London topics (see Appendix B). Each question was repeated 25 times per topic (or knowledge base), yielding 500 responses per knowledge base and 1,000 responses in total.

\section{Results}

For each response from the experiments, we did the following.

\begin{itemize}
\item Used a standard system prompt: "You are a helpful assistant."
\item Recorded the complete text response
\item Generated an embedding vector using the \texttt{gte-Qwen2-1.5B-instruct} embedding model [23]
\end{itemize}

We then analyzed the responses in the embedding vector space.

\subsection{Distinguishing different models}

\begin{figure}[ht]
\centering
\fbox{\includegraphics[width=\linewidth]{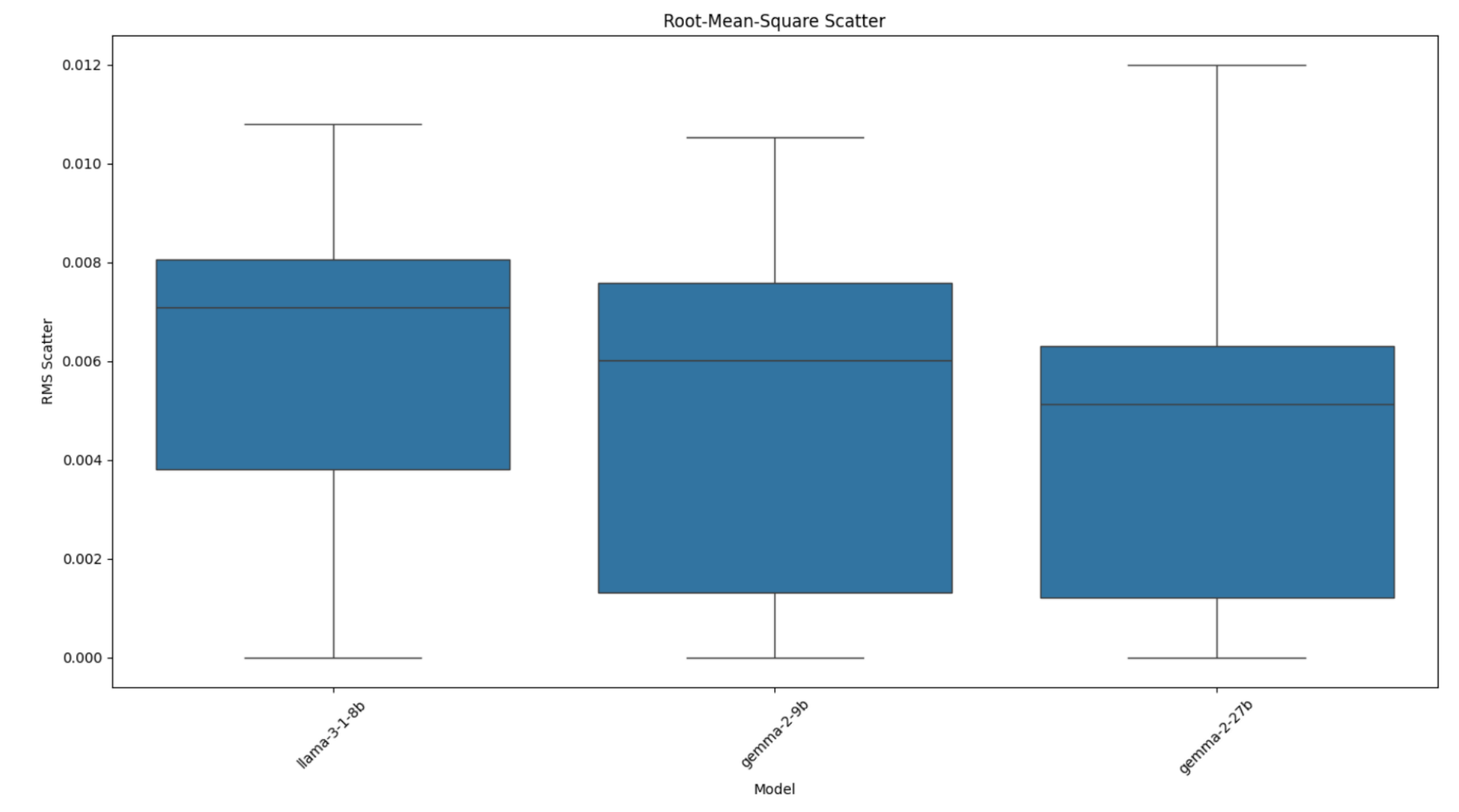}}
\caption{Internal consistency of different LLMs measured by RMS scatter.}
\label{fig:llm-consistency}
\end{figure}

Figure \ref{fig:llm-consistency} shows the internal consistency of answers from the same question for each LLM, measured by the RMS scatter.

The consistency metrics reveal that Gemma-2-27b demonstrates the highest consistency in responses, with an RMS scatter of $0.0043$. Llama-3.1-8b shows the highest variation with an RMS scatter of $0.0062$.

\begin{figure}[ht]
\centering
\fbox{\includegraphics[width=\linewidth]{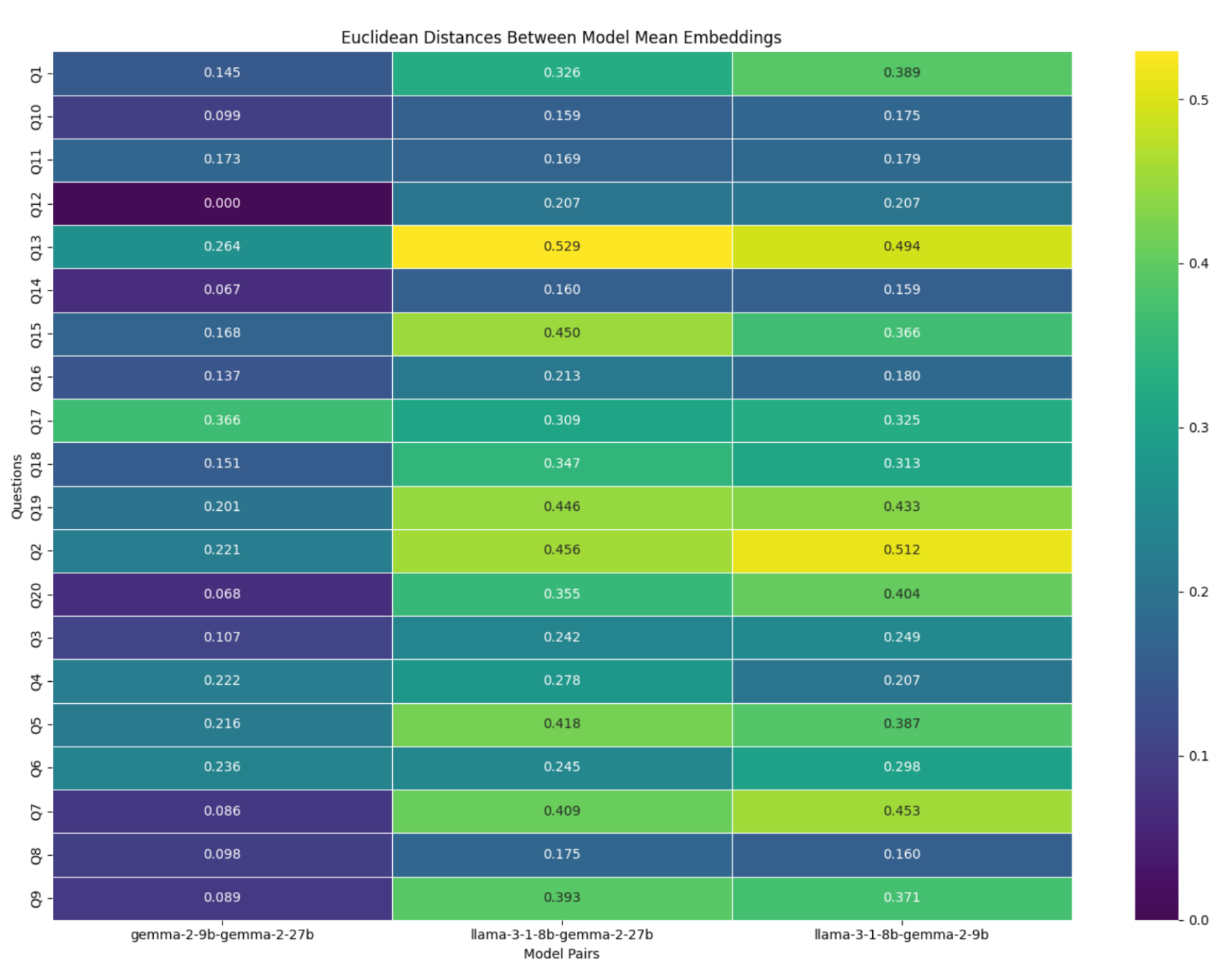}}
\caption{Average distance vs RMS scatter for each question and model pair.}
\label{fig:llm-distance}
\end{figure}

To evaluate whether different models give different answers for the same question, Figure \ref{fig:llm-distance} shows the mean distance of answer clusters generated by pairs of LLMs.

The distance matrix reveals that inter-model distances vary substantially across questions. For instance, the highest distance observed was $0.5291$ between Llama-3.1-8b and Gemma-2-27b for question Q13 ("Who wrote 'Romeo and Juliet'?"). The lowest non-zero distance was $0.0669$ between Gemma models for question Q14 ("What is the atomic number of oxygen?").

Critically, Table \ref{tab:llm-ratio} shows that the distances between model pairs are $32-65$ times larger than the highest RMS scatter observed within any model ($0.0062$ for Llama-3.1-8b). This large separation between inter-model distances and intra-model variation indicates that Gaia nodes with different LLMs produce reliably distinguishable outputs, making them identifiable through their response patterns.

\begin{table}[ht]
\centering
\caption{Average distance $D_{ave}$ over max RMS scatter $\sigma_{max}$ for each pair of LLMs}
\begin{tabular}{ccc}
\hline
Model pair & $D_{ave}$ & $D_{ave}/\sigma_{max}$ \\
\hline
Gemma9b - Gemma27b & $0.1558$ & $32.5\times$ \\
Llama8b - Gemma9b & $0.3129$ & $65.2\times$ \\
Llama8b - Gemma27b & $0.3141$ & $65.4\times$ \\
\hline
\end{tabular}
\label{tab:llm-ratio}
\end{table}

\subsection{Distinguishing different knowledge bases}

\begin{figure}[ht]
\centering
\fbox{\includegraphics[width=\linewidth]{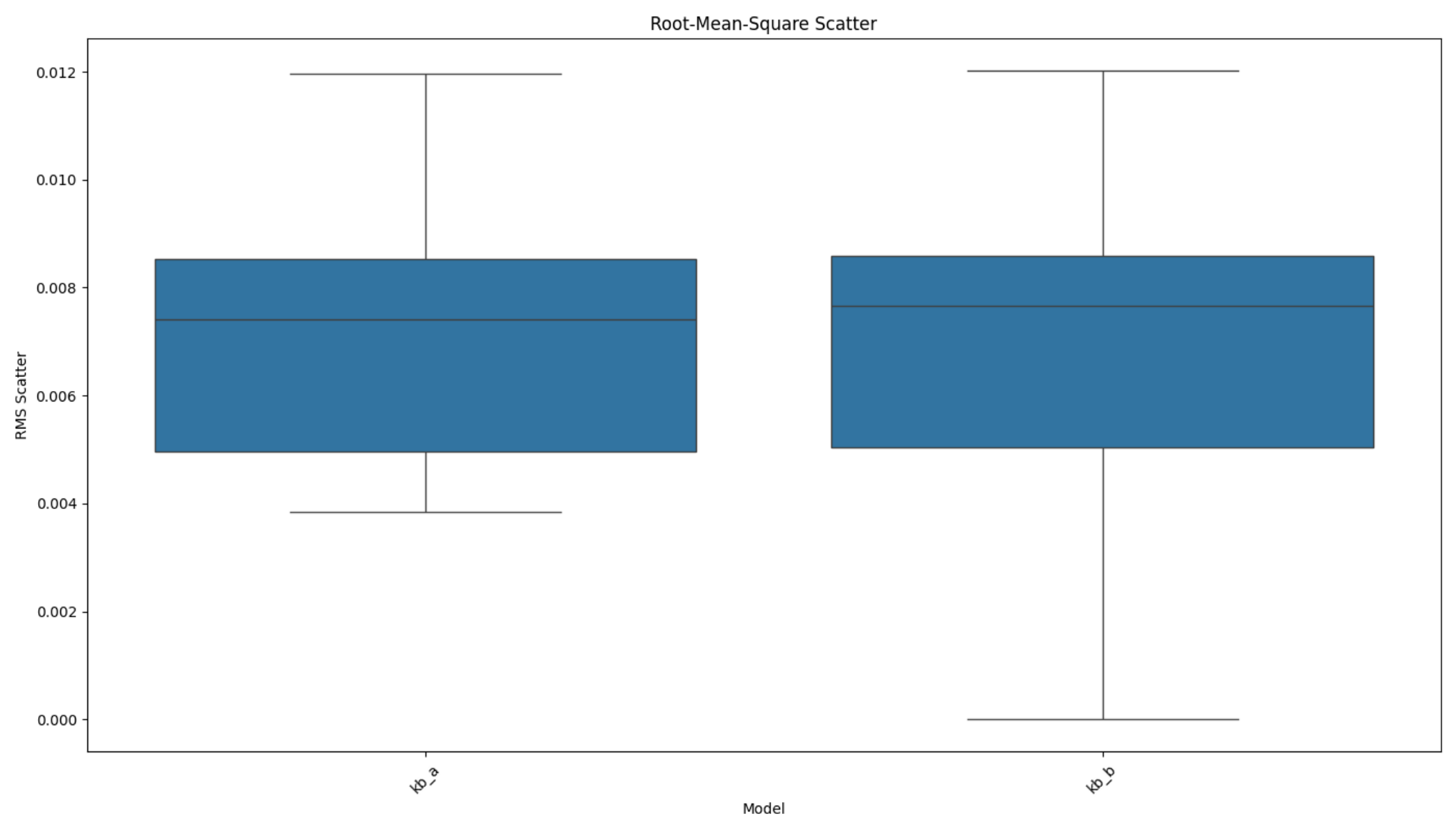}}
\caption{Internal consistency of different knowledge bases measured by RMS scatter.}
\label{fig:kb-consistency}
\end{figure}

Figure \ref{fig:kb-consistency} shows the internal consistency of answers from the same question for each knowledge base, measured by the RMS scatter. The Paris knowledge base is denoted as \texttt{kb\_a}, and the London knowledge base is denoted as \texttt{kb\_b} in the figures.

\begin{figure}[ht]
\centering
\fbox{\includegraphics[width=\linewidth]{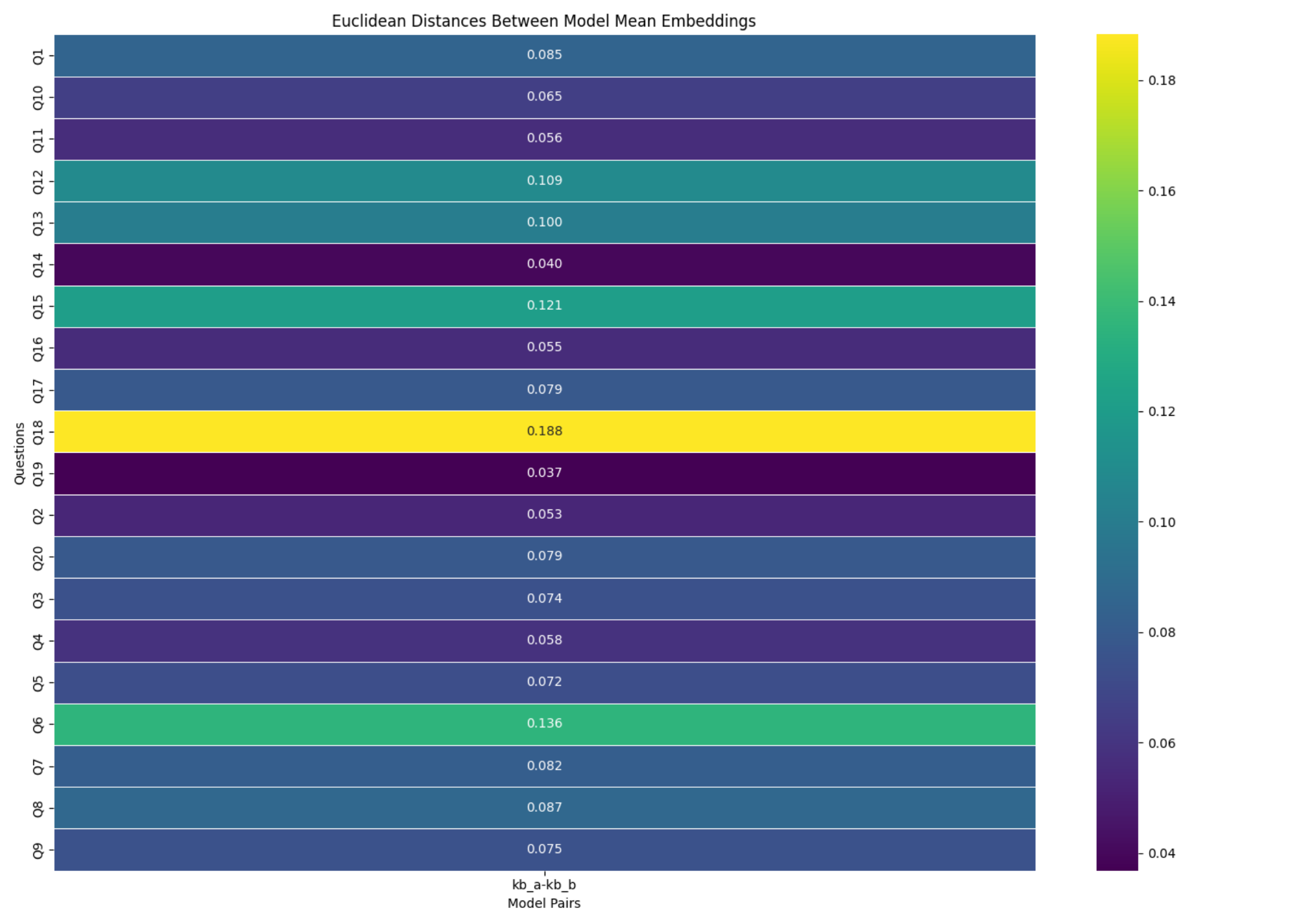}}
\caption{Average distance vs RMS scatter for each question and knowledge base pair.}
\label{fig:kb-distance}
\end{figure}

The consistency metrics show that the two knowledge bases generate responses of similar consistency. To evaluate whether nodes with different knowledge bases give different answers for the same question, Figure \ref{fig:kb-distance} shows the mean distance of answer clusters generated by the two knowledge bases.

The distance matrix reveals that inter-knowledge base distances vary substantially across questions. For instance, the highest distance observed was $0.188$ between Paris and London knowledge bases for question Q18 ("How many bridges did Philip Augustus build in Paris in the late 12th century?"). The lowest non-zero distance was $0.037$ between knowledge bases for question Q19 ("What percentage of Paris's salaried employees work in hotels and restaurants according to the document?").

Critically, the distances between knowledge base pairs (average $0.0862$) are about $5-26$ times larger than the RMS scatter observed within a knowledge base ($0.0072$). This large separation between inter-knowledge base distances and intra-knowledge base variation indicates that Gaia nodes with different knowledge bases produce reliably distinguishable outputs, making them identifiable through their response patterns.

Our hypotheses are validated by the experimental results.

\section{Discussions}

The ratios between inter-model / inter-knowledge base distances and intra-model / intra-knowledge base variations are statistically substantial. Further research would be needed to determine how this ratio might be affected by factors such as changing hardware, varying load conditions, or minor model updates. These initial results suggest a potential basis for statistical differentiation between models and knowledge bases.

The two Gemma models are more similar to each other (distance of $0.1558$) than either is to Llama (distances around $0.31$). While still $32\times$ greater than the observed internal variation, this suggests that distinguishing between models within the same family might present different challenges than distinguishing between models from different families.

Paradoxically, different knowledge bases produce more similar answers (distances around $0.1$) than different LLM models. The reason is probably due to that our two knowledge bases are both about European capitals. While the answers about London or Paris are meaningfully different, they might actually look similar to the embedding models. That said, the inter-knowledge base distances are still substantially larger than scatters within a knowledge base, allowing observers to distinguish between knowledge bases from statistical properties of the answers.

Finally, different questions demonstrated varying levels of differentiation between models or knowledge bases. For example, Q13 ("Who wrote 'Romeo and Juliet'?") showed the greatest distance between models ($0.5291$ between Llama 8b and Gemma 27b). Q18 ("How many bridges did Philip Augustus build in Paris in the late 12th century?") showed the greatest distance between knowledge bases ($0.188$ between London and Paris). Those variations in the effectiveness of test questions are important in practical applications. We will design a system that can identify and rank the most effective test questions through real-world use cases. 

\section{System design and implementation}

Gaia domains are permissionless collections of AI inference nodes (Gaia node). Each Gaia node in a Gaia domain agrees to run the exact LLM (GGUF model file) and knowledge base (Qdrant vector snapshot) specified by the domain owner. That allows the Gaia domain, as a whole, to provide consistent services to the public. 

In this paper, we propose an EigenLayer AVS (Actively Verified Services) design that is tailored to the Gaia network architecture. The AVS is organized into multiple "operator sets". 

\begin{itemize}
\item The operator set 0 is a special group for AVS validators. An AVS validator is responsible for polling Gaia nodes with questions and performing statistical analysis to detect outliers within each Gaia domain. In order to join operator set 0, the validator must be approved by the Gaia DAO organization.
\item The operator sets 1 to $n$ are mapped to Gaia domains. All Gaia nodes in each domain form a single operator set.
\end{itemize}

The AVS operates on Epochs. Each epoch lasts 12 hours. During the epoch, each validator would perform the following tasks repeatedly.

\begin{itemize}
\item Poll each Gaia node in a domain using a random question from a question set that is specially designed for this domain.
\item Record all responses, time-outs, and error messages.
\item Detect outlier nodes based on responses.
\item Create time-encrypted and signed messages for the results.
\item Post the results on EigenDA [24].
\end{itemize}

The initial question set will be provided by the domain owner. But as the AVS runs over time, the validators will collect and filter real-world user questions that are most effective in generating tightly clustered answers from the majority of operator nodes.

At the end of each Epoch, a randomly chosen validator will act as the aggregator and perform a map-reduce function on all messages posted by all validators in that Epoch. For applications in regulated industries, it is possible to utilize Secure Multi-Party Computation (SMPC) methods so that confidential data is not disclosed to validators and the aggregator. The result is a set of flags for each Gaia node, such as 

\begin{itemize}
\item \texttt{outlier} -- The Gaia node produces responses that are statistical outliers compared to other nodes in the same domain, indicating that it might be running the incorrect LLM or knowledge base required for the domain.
\item \texttt{slow} -- The Gaia node responds much slower (i.e., outside of three standard deviations from the mean response time) than nodes in the same domain.
\item \texttt{timeout} -- The Gaia node timed out in one or more requests.
\item \texttt{error 500} -- The Gaia node returned HTTP error code 500, indicating an internal server error, in one or more requests.
\item \texttt{error 404} -- The Gaia node returned HTTP error code 404, indicating unavailable resources, in one or more requests.
\item \texttt{error other} -- The Gaia node returned any other HTTP error code in one or more requests.
\end{itemize}

The cumulative status of a Gaia node (i.e., AVS operator) at the end of each Epoch determines its AVS award or punishment from this Epoch. For example, if a Gaia node maintains a good status, without any flags, in the past 10 epochs, it will receive the regular AVS award. If it is flagged as an \texttt{outlier} or an \texttt{error 500} in the current epoch, or flagged as \texttt{slow} for several epochs in a row, it might be suspended from both the AVS award and domain participation for the next several epochs. For malicious actors, the AVS might even slash their stakes. The detailed rules on awards and punishments will be determined by the Gaia network.

Furthermore, participating domains could also ask the AVS validators to onboard new Gaia nodes. Currently, a Gaia node can apply to join a domain. The domain owner reviews and approves the application. The AVS validators could take over this role. They could poll the candidate node with requests specific to the domain. Gaia nodes that conform to the domain requirements for LLM, knowledge base, and response speed will be automatically admitted as the validator results are posted in EigenDA.

\section{Cryptoeconomic considerations}

The AVS validators detect outlier nodes that provide "abnormal" answers to the test questions. The approach is statistical and non-deterministic. There could be errors in both type I (false positive, in which an honest node is detected incorrectly) and type II (false negative, in which an dishonest node is not detected). Hence, the "punishment" of supposedly "dishonest" nodes should be gradual and reversible. For exmaple,

\begin{itemize}

\item We should have strict slashing rules and conditions. First-time outliers should not be slashed. The AVS could suspend them so that they do not receive any revenue from AI applications and do not receive staking awards.

\item Repeated outliers and offenders could get slashed. But depending on the severity of the violation, the validator could slash the partial or the full amount of the stake.

\end{itemize}

The AVS validators are also staked. If a validator consistently disagrees with the majority of validators, he could also be suspended or even slashed.

It is possible for the cryptoeconomic model to fail if the AVS staked value becomes too low. In that case, the outlier nodes would receive very light "punishment" for their dishonest behavior. On the other hand, requiring a large stake would increase the barrier of entry for node operators, which makes the system less secure and less capable by reducing the diversity across node operators.

It is also possible for Gaia nodes in a domain to collude. They could all run an AI model that is different from the domain's specification. In that case, the AVS will force all nodes in the domain to run the majority-chosen AI model. The AVS will remain self-consistent but it could mis-match the domain's public advertisement or declaration. Users will find poor service quality from that Gaia domain, and will switch to alternative Gaia domains. The free market will ensure that the honest and healthy domains (i.e., free of collusion) will capture the most economic value.

Each AVS operator set (i.e., Gaia domain) can only scale to 1000 nodes at the time of this writing. There is no inherent limit on how many operators a Gaia validator can check. The Gaia AVS validator could broadcast the test question to all its operator nodes in one go and receive responses asynchronously. The 1000-node limit is a technical limitation for the validator nodes to form consensus in Eigen's suite of smart contracts. As Eigen improves the efficiency of its system, we expect that the limit will be drastically increased or even removed.

Finally, other AVS approaches have been suggested to support verifiable agents through on-chain smart contracts. However, the current computational limits for on-chain smart contracts make it impossible to run LLM inference on-chain. Eigen's Layer 1 Agents proposal [25] focuses on verifiable LLM tools, such as function calls and memory use. It calls for LLM tools to be hosted on their own AVSes so that we can ensure that LLMs are getting verifiable data and performing verifiable actions. Those AVSes for LLM tools are highly complementary to the LLM inference AVSes discussed in this paper.

\section{Conclusions}

In this paper, we demonstrated that statistical analysis of LLM outputs could reliably signal the model and knowledge base in use. That allows decentralized LLM inference networks to use EigenLayer AVS to verify LLM outputs intersubjectively and detect outliers as potential bad actors. Combined with cryptoeconomic incentives and penalties in AVSes, it is possible to scale this LLM verification approach to large networks. 

\section{References}

\begin{flushleft}
[1] Zhao, W. X. et al. A Survey of Large Language Models, \url{https://arxiv.org/abs/2303.18223}, v16, 2025

\smallskip
[2] Fourrier, C., Habib, N., Lozovskaya, A., Szafer, K., Wolf, T. Performances are plateauing, let's make the leaderboard steep again, \url{https://huggingface.co/spaces/open-llm-leaderboard/blog}, 2024

\smallskip
[3] Jiang, J., Wang, F., Shen, J., Kim, S., Kim S. A Survey on Large Language Models for Code Generation, \url{https://arxiv.org/abs/2406.00515}, 2024

\smallskip
[4] Gaia Foundation. GaiaNet: GenAI Agent Network, \url{https://docs.gaianet.ai/litepaper}, 2024

\smallskip
[5] Goldwasser, S., Micali, S., Rackoff, C. The knowledge complexity of interactive proof-systems, Proceedings of the seventeenth annual ACM symposium on Theory of computing, 1985

\smallskip
[6] Sun, H., Li, J., Zhang, H. zkLLM: Zero Knowledge Proofs for Large Language Models, \url{https://arxiv.org/abs/2404.16109}, 2024

\smallskip
[7] Ganescu, B-M., Passerat-Palmbach, J. Trust the Process: Zero-Knowledge Machine Learning to Enhance Trust in Generative AI Interactions, The 5th AAAI Workshop on Privacy-Preserving Artificial Intelligence, 2024

\smallskip
[8] Dong, B., Wang, Q. Evaluating the Performance of the DeepSeek Model in Confidential Computing Environment, \url{https://arxiv.org/abs/2502.11347v1}, 2025

\smallskip
[9] Apsey, E., Rogers, P., O'Connor, M. \& Nertney, R. Confidential Computing on NVIDIA H100 GPUs for Secure and Trustworthy AI, \url{https://tinyurl.com/9henzz7t}, 2023

\smallskip
[10] Oasis. A Cup of TEE, Please. But How Do We Know It's The Right Flavor?, \url{https://oasisprotocol.org/blog/tees-remote-attestation-process}, 2025

\smallskip
[11] Karpukhin, V., Oguz, B., Min, S., Lewis, P., Wu, L., Edunov, S., Chen, D. and Yih, W. Dense passage retrieval for open-domain question answering. In EMNLP (1), pp. 6769â€"6781, 2020.

\smallskip
[12] EigenLayer. EigenLayer: The Restaking Collective, \url{https://docs.eigenlayer.xyz/assets/files/EigenLayer_WhitePaper-88c47923ca0319870c611decd6e562ad.pdf}, 2023

\smallskip
[13] EigenLayer. EIGEN: The Universal Intersubjective Work Token, \url{https://docs.eigenlayer.xyz/assets/files/EIGEN_Token_Whitepaper-0df8e17b7efa052fd2a22e1ade9c6f69.pdf}, 2024

\smallskip
[14] Gaia Foundation. Verifiable Inference, \url{https://github.com/GaiaNet-AI/verifiable-inference}, 2025a

\smallskip
[15] Grattafiori, A. et al. The Llama 3 Herd of Models, \url{https://arxiv.org/abs/2407.21783}, 2024

\smallskip
[16] Mesnard, T. et al. Gemma: Open Models Based on Gemini Research and Technology, \url{https://arxiv.org/abs/2403.08295}, 2024

\smallskip
[17] Gaia Foundation. Gaia nodes with long-term knowledge, \url{https://docs.gaianet.ai/tutorial/concepts}, 2025b

\smallskip
[18] Wikipedia. Paris, \url{https://en.wikipedia.org/wiki/Paris}, 2025a

\smallskip
[19] Wikipedia. London, \url{https://en.wikipedia.org/wiki/London}, 2025b

\smallskip
[20] Nussbaum, Z., Morris, J. X., Duderstadt, B. and Mulyar, A. Nomic Embed: Training a Reproducible Long Context Text Embedder, \url{https://arxiv.org/abs/2402.01613}, 2025

\smallskip
[21] Yuan, J. Paris knowledge base, \url{https://huggingface.co/datasets/gaianet/paris}, 2024a

\smallskip
[22] Yuan, J. London knowledge base, \url{https://huggingface.co/datasets/gaianet/london}, 2024b

\smallskip
[23] Yang, A. et al. Qwen2 Technical Report, \url{https://arxiv.org/abs/2407.10671}, 2024

\smallskip
[24] EigenDA. What is EigenDA? \url{https://docs.eigenda.xyz/core-concepts/overview}, 2025

\smallskip
[25] EigenLayer. Introducing Verifiable Agents on EigenLayer, \url{https://blog.eigencloud.xyz/introducing-verifiable-agents-on-eigenlayer/}, 2025

\end{flushleft}


\section{Appendix A: Sample questions for distinguishing models}

\begin{enumerate}
\item What year did the Apollo 11 mission land on the moon?
\item Who wrote Pride and Prejudice?
\item What is the capital of Japan?
\item What is the chemical formula for water?
\item Who painted the Mona Lisa?
\item What is the largest planet in our solar system?
\item In what year did World War II end?
\item What is the speed of light in a vacuum?
\item Who discovered penicillin?
\item What is the boiling point of water at sea level in Celsius?
\item Who was the first woman to win a Nobel Prize?
\item What is the capital of Australia?
\item Who wrote 'Romeo and Juliet'?
\item What is the atomic number of oxygen?
\item Who was the first person to step on the moon?
\item What is the circumference of Earth?
\item In which year was the Declaration of Independence signed?
\item What is the tallest mountain on Earth?
\item Who developed the theory of relativity?
\item What is the largest ocean on Earth?
\end{enumerate}

\section{Appendix B: Sample questions for distinguishing knowledge bases}

\begin{enumerate}
\item What was the population of Greater London according to the 2011 census, and what was the population density per square mile?
\item In which year did the Romans found Londinium, and what event in 61 AD led to its destruction?
\item Who supervised the rebuilding of London after the Great Fire of 1666, and which architect's cathedral was completed in 1708?
\item What are the four World Heritage Sites in London, including the combined site that consists of three connected landmarks?
\item When was Greater London divided into 32 London boroughs plus the City of London, and which administrative body was abolished in 1889?
\item What was the name of the defensive perimeter wall built during the English Civil War, how many people were involved in building it, and when was it leveled?
\item What is the highest temperature ever recorded in London, on what date did it occur, and at which measuring station?
\item Which two areas are London's main financial districts, and which one has recently developed into a financial hub?
\item According to the 2021 census data, what percentage of London's population was foreign-born, and what were the five largest countries of origin?
\item What governmental body is responsible for London's transport system, and what is the name of the functional arm through which it operates?
\item How many visitors did the Louvre Museum receive in 2023?
\item What is the name of the prefect who supervised the massive public works project that rebuilt Paris between 1853 and 1870?
\item What happened to Paris during the Fronde civil war in the 17th century?
\item What is the unemployment rate in Paris as reported in the 4th trimester of 2021?
\item What percentage of Parisians earned less than â‚¬977 per month in 2012, according to the text?
\item Which organization created the VÃ©lib' bicycle sharing system in Paris and in what year?
\item What is the name of the Paris stock exchange mentioned in the document?
\item How many bridges did Philip Augustus build in Paris in the late 12th century?
\item What percentage of Paris's salaried employees work in hotels and restaurants according to the document?
\item What was the date when Paris was liberated from German occupation during World War II?
\end{enumerate}

\end{document}